# New Results for the Text Recognition of Arabic Maghribī Manuscripts - Managing an Under-resourced Script

Noëmie Lucas (Islamic and Middle Eastern Studies, University of Edinburgh)
Clément Salah (Sorbonne Université – Institut d'Histoire et Anthropologie des Religions, Faculté de Théologie et Sciences des Religions, Université de Lausanne)
Chahan Vidal-Gorène (Ecole Nationale des Chartes, Université Paris Sciences & Lettres – Calfa)

**Abstract**: HTR models development has become a conventional step for digital humanities projects. The performance of these models, often quite high, relies on manual transcription and numerous handwritten documents. Although the method has proven successful for Latin scripts, a similar amount of data is not yet achievable for scripts considered poorly-endowed, like Arabic scripts. In that respect, we are introducing and assessing a new *modus operandi* for HTR models development and fine-tuning dedicated to the Arabic Maghribī scripts. The comparison between several state-of-the-art HTR demonstrates the relevance of a word-based neural approach specialized for Arabic, capable to achieve an error rate below 5% with only 10 pages manually transcribed. These results open new perspectives for Arabic scripts processing and more generally for poorly-endowed languages processing. This research is part of the development of RASAM dataset in partnership with the GIS MOMM and the BULAC.

# Background

What does it mean to edit a manuscript in the digital age? What does that change in terms of codicology, paleography, and philology? What is digital philology in comparison with old philology? Cornelis Van Lit recently offered the first handbook to guide students and scholars in that path with his work *Among Digitized Manuscripts. Philology, Codicology, Paleography in the Digital World.* (van Lit [2019]) In his work, he precisely ambitions to help scholars to take advantage of computer power and computational methods in their study of manuscripts. The study of manuscripts and literary texts refers to philology and includes the edition of the text, the textual criticism, the history of the text as a book, or the history of its transmission. This ancient discipline has known somewhat of a revival thanks to the use of computers and the implementation of calculation methods. Therefore, we tend to talk about "Digital philology" or "Computational philology" to pertain to a discipline at the crossroads between classical sciences of the text and computational methods. This philology is data-oriented philology. Its specificity lies in its "respective willingness to divide labor between human and artificial intelligence." (Andrews [2013]: 62)

When it comes to data production, or text edition, the digital philology can be applied at each level: the acquisition of the text (HTR), the structuration of the text (encoding), enrichment and indexing (lemmatization for example). This article addresses the question of the acquisition of the text, and particularly one way to acquire a text, which consists of a computer-assisted transcription using an algorithm for optical character recognition (OCR) or for handwriting text recognition (HTR). We will mainly deal with *handwritten text recognition* since we are working on manuscripts copied between the ninth and the nineteenth centuries.

# I. The Arabic manuscripts in the digital age

## 1. HTR and Arabic manuscripts

OCR or HTR refers to automatic text recognition software that analyses a scanned image to extract the text. A distinction is commonly made between OCR and HTR; the first applies to printed characters, the second to handwritten texts. The difference between the two is rooted in the different technical challenges they face, especially regarding the layout, and the related results. If on one hand OCR is now considered to be a resolved issue, at least for Latin languages (character error rate or CER less than 2%) with free or commercial software, HTR, on the other hand, has long remained largely non-functional because the challenges are more complex, due to the high variability of documents and scripts. The first HTR tests date back to the 1980s, but it is thanks to the development of artificial intelligence and neural networks over the last ten years that things have been able to improve. This is a very dynamic field of research, as there exists no general model, even for Latin languages. The training of specialized models on a type of writing, a type of hand, or a type of manuscript, can easily reach a CER of about 5%, or even less in the case of a simple layout (Neudecker, *et al.* [2019]).

In a few words, the deep learning method consists in submitting to a neural network databases containing multiple samples of what we wish the network to recognize. The network extracts information from these databases and learns to generalize the object or text and to recognize it regardless of the context. Training is thus achieved through frequency and habit.

OCR or HTR technologies are not the relegation of the text transcription task to artificial intelligence. It is an interactive process between human and machine. For the machine to be able to predict the content of a text, it will first have to rely on a transcription phase conducted manually followed by an automated calculation phase. The next stage will consist of a human correction phase followed by a re-training phase, until a satisfactory model is reached.

OCR and HTR systems have been developed for Arabic scripts in historical documents for a few years now. Among the initiatives and projects, one cannot ignore the work undertaken by the OpenITI project and, first, by Benjamin Kiessling regarding the OCR of Arabic printed editions (Romanov, *et al* [2017]), and more recently with two important projects the *Open Islamicate Text Initiative Arabic-script OCR Catalyst Project* devoted to Arabic OCR and the ACDC project, which stands for *Automatic Collation for Diversifying Corpora*.[1] The latter project aims at improving HTR for Persian and Arabic manuscripts and focuses on creating training data for developing generalizable models. Since 2020, they are also collaborating with the team of eScriptorium, one of the latest platforms for Historical Document Analysis, which focuses on manual and automatic transcription (using Kraken), image, and textual annotation.

Other research has been carried out in line with OCR and HTR of Arabic scripts like the *Automatic Transcription of Historical Handwritten Arabic Texts* project of the British Library,[2] the research of Thomas Milo and Alicia Gonzalez (González Martínez, Milo [2020]), or the work done by David Wrisley and Süphan Kirmizialtin at NYU Abu Dhabi on Ottoman periodicals (Wrisley, Kirmizialtin [2020]). The British Library team and also NYU Abu Dhabi's notably use a well-known platform called Transkribus, which has been developed in the framework of the READ project in Austria since 2013 (now commercial). Alongside turn-key OCR for Arabic scripts like ABBYY (commercial) and Tesseract (open-source and hence included in several other OCR webapps), many platforms, free or not, open-source or not, exist to transcribe documents with the help of artificial intelligence, such as OCR4all, eScriptorium, Calfa Vision or Transkribus.[3] A recent contribution written by Ishida Yuri and Shinoda Tomoaki assesses how accurate for Arabic were a dozen OCR systems known for being easy to use and not too expensive.[4] Among the eleven systems tested for their abilities to process Arabic,[5] i2 OCR, OCR Space, Google Drive (that being Tesseract) and Fine Reader (ABBYY) were the most robust options with a CER between 5% and 10%.[6] It is worth mentioning that they were evaluating OCR systems and not HTR systems, hence the absence of some important platforms for handwriting recognition like Transkribus.

All in all, HTR research for Arabic scripts is in progress albeit numerous issues encountered, due to non-Latin languages characteristics notably. In comparison with other Latin scripts, Arabic writings do not benefit from the same amount of datasets and training data although they are key to developing strong and versatile models for a language.
The research we will present in this article is in line with past and ongoing research. It aims to contribute to the development of HTR for Arabic handwritten scripts and to identify data needs for such an under-resourced script. A language or a script may still be regarded as under-resourced on several accounts:
- **the data accessibility** is still an impediment to the creation of systems, which usually rely on a massive amount of data for efficiency;

---

[1] To learn more about recent developments, M. Miller and S. Savant, "OpenITI OCR and Text Production teleconference," Online presentation, April 19, 2022: https://www.youtube.com/watch?v=Cb5FI9NUJXY

[2] https://www.bl.uk/projects/arabic-htr

[3] https://vision.calfa.fr; https://www.escriptorium.fr; https://readcoop.eu/transkribus; https://www.ocr4all.org/

[4] https://digitalorientalist.com/2021/09/17/a-study-on-the-accuracy-of-low-cost-user-friendly-ocr-systems-for-arabic-part-1/
https://digitalorientalist.com/2021/09/24/a-study-on-the-accuracy-of-low-cost-user-friendly-ocr-systems-for-arabic-part-2/

[5] Convertio (https://convertio.co/); ABBYY Fine Reader PDF (https://pdf.abbyy.com/pricing/)
Foxit Phantom PDF, Free Online OCR (https://www.newocr.com/); Gold/Sakhr (http://www.sakhr.com/index.php/en/solutions/ocr); i2 OCR (https://www.i2ocr.com/free-online-arabic-ocr); OCR Convert (https://www.ocrconvert.com/arabic-ocr); OCR Space (https://ocr.space/); Online Convert Free (https://onlineconvertfree.com/ocr/arabic/); Sotoor (https://rdi-eg.ai/optical-character-recognition/).

[6] More details in the blog post (cf. note 4.)

- **the number of specialists** able to transcribe and annotate quickly these data in order to have a sufficient dataset for HTR models training;
- the less adequate approaches for non-Latin scripts arising from the current technologies **specialization for Latin scripts**.

Besides, there is also the requirement for machine learning knowledge in order to harness the OCR and HTR systems to their full potential.

## 2. RASAM and Calfa Vision

The creation of a dataset for the HTR of Arabic manuscripts from the Maghrib was the goal of a partnership between the French Research Consortium "Middle East and Muslim Worlds", the Bibliothèque Universitaire des Langues et Civilisations (BULAC) and Calfa[7] in 2021. It indeed resulted in the creation of a dataset called RASAM and enabled the training of a HTR model for the Arabic scripts (Vidal-Gorène, *et al* [2021b]). One of the objectives of the hackathon for the collaborative transcription of manuscripts was scientific and experimental. We aimed to assess the feasibility of automatic text recognition for one type of manuscripts, Arabic Maghribī script manuscripts. On top of that, we ambitioned to build and provide to the research community a solid dataset for recognition of Arabic Maghribī scripts that extends existing datasets such as BADAM or RASM for layout analysis.[8] The dataset created is the achievement of a collaborative work; it comprises 300 images, with their related ground truth stored in a XML file offering three levels of information: text regions with their coordinates and semantic tag, text lines with the coordinates and transcription and baselines coordinates.[9]

These images come from three manuscripts (MS.ARA.609, MS.ARA.417 and MS.ARA.1977)[10] preserved at the BULAC and available on the online library of the BULAC (BINA)[11]. Several experiments have been made and numerous models have been trained in the course of this project, and the final HTR model reached a character error rate of 4.8%, meaning that more than 95% characters were well predicted (Vidal-Gorène, *et al* [2021b]: 279). The RASAM model has been developed from a limited number of manuscripts in order to obtain a "pre-generical" model for Arabic Maghribī writings, which will demonstrate a high strength and be a solid start to train new models. The research and experiments that we are presenting in this paper are in line with this project and intend to extend it. This paper addresses the versatility of the RASAM dataset and evaluates its practical use for the analysis of other manuscripts in Maghribī scripts.

---

[7] Calfa is a company specialized in document understanding and text recognition for oriental languages. The company is involved in the digitization of printed and handwritten materials in Armenian, Georgian, Syriac and Arabic scripts in particular.

[8] BADAM, see B. Kiessling, D. Stökl Ben Ezra, M.T. Miller, "BADAM: A Public Dataset for Baseline Detection in Arabic-script Manuscripts," in *Proceedings of the 5th International Workshop on Historical Document Imaging and Processing*. 2019. 13-18 ; RASM, see C. Clausner, A. Antonacopoulos, N. Mcgregor and D. Wilson-Nunn, "ICFHR 2018 Competition on Recognition of Historical Arabic Scientific Manuscripts - RASM2018," in *16th International Conference on Frontiers in Handwriting Recognition (ICFHR)*, Niagara Falls, (NY, USA, 2018), 471-476. https://doi:10.1109/ICFHR-2018.2018.00088 ; Datasets available: BADAM: https://zenodo.org/record/3274428#.YW_xJhrMI2w RASM: https://bl.iro.bl.uk/concern/datasets/f866aefa-b025-4675-b37d-44647649ba71?locale=en

[9] RASAM dataset available here: https://github.com/calfa-co/rasam-dataset

[10] MS.ARA.609: https://bina.bulac.fr/ARA/MS.ARA.609; MS.ARA.417: https://bina.bulac.fr/ARA/MS.ARA.417; MS.ARA.1977: https://bina.bulac.fr/ARA/MS.ARA.1977

[11] https://bina.bulac.fr/

For RASAM, as well as for the present research, the creation of training data and the development of a HTR model specialized in one type of Arabic script has been then realized with the Calfa Vision platform. This platform has been developed by Calfa in order to overcome these specific issues of data and model creation, with a focus on oriental scripts, that are mainly not covered by the existing systems for the general public or for researchers. The platform was originally created for the specific processing of Armenian (language with a complex abbreviative system, notably with the use of ideograms) and was gradually extended to the specific processing of Semitic languages (especially Syriac) and other oriental scripts languages (Greek, Georgian, etc.). Several approaches for HTR can be defined depending on the language considered or the requirements specifications. Whereas a character-based approach seems appropriate for ancient Armenian or Latin manuscripts, the word separation is proving to be difficult for *scriptio continua* or a highly abbreviated text. A new approach, word-based, is proving more appropriate, with a 7% decrease of the error rate at the character level (Camps, *et al* [2021]). Thus, the collaborative platform incorporates pre-trained models for automatic document analysis (layout and HTR), that are strengthened through checking and transcription of new documents within a project in an iterative way, in order to achieve quickly specialized models dedicated to the edition issues at hand.[12]

## 3. HTR as a way to valorize Arabic Maghribī script manuscript collections

The RASAM project, as well as the present research, envision other purposes alongside technical and experimental objectives: the valorization of Maghribī manuscript collections and the promotion of studies on the Maghrib (either medieval or modern ones). Collections of Arabic Maghribī scripts manuscripts are well represented mainly in Spanish and Maghribī countries libraries, which is not surprising knowing that manuscript tradition in the Maghrib and Andalusia is a long lived one.[13] We also find numerous collections of Maghribī script manuscripts in "Oriental" libraries like the Suleymaniye of Istanbul and in countries like France, due to its colonial history notably.[14] It is fair to say however that these collections of manuscripts are under-exploited compared to the collection of Oriental manuscripts. One of the reasons for this could be that fewer parts of these manuscripts are digitized, and even when digitized are not necessarily available online.[15] To give but one example of the digitization policies with regard to these collections, we can mention what Morocco does. Morocco is known to have supported these past few years some digitization programs. In some cases, these projects aim at spreading Moroccan heritage to a larger public and to the international scientific community like what the Bibliothèque Nationale du Royaume du Maroc (BNRM) in Rabat[16] did or with the work done by the Fondation du Roi Abdul-Aziz Al Saoud in Casablanca.[17] In other cases, like the Royal library Ḥassaniya, digitization is seen as a way to preserve the manuscripts and does not come with an uploading of these digitized manuscripts online (the Ḥassaniya for instance does not even have a website). Our ongoing research about HTR and Arabic Maghribī manuscripts intends to offer accurate tools to allow broader projects on these manuscripts, whose importance and historical richness for the history of the Maghrib is without contest.

---

[12] For a detailed panel of functionalities and examples of model specialization on a given task, see: C. Vidal-Gorène, B. Dupin, A. Decours-Perez, and T. Riccioli. "A modular and automated annotation platform for handwritings: evaluation on under-resourced languages" in *International Conference on Document Analysis and Recognition*, (Springer: Cham, 2021), 507–522.

[13] Among the main libraries, we can mention the Bibliothèque Nationale du Royaume du Maroc in Rabat (Morocco), the Bibliothèque Nationale d'Algérie in Alger (Algeria), the Bibliothèque Nationale de Tunisie in Tunis (Tunisia), or the Escurial at Madrid (Spain).

[14] In France, the Bibliothèque Nationale de France (BNF) and the Bibliothèque Universitaire des LAngues et Civilisation (BULAC) in Paris kept numerous Maghribī manuscripts. The BULAC, for example, holds the second largest collection of Arabic manuscripts in France. In May 2022, 33% of the collection has been reported and cataloged in online repository (the collection comprises about 2,500 documentary units), that represents 814

A very dynamic field, the Maghribī studies have seen a revival in the past decades in relation to the Arab Springs, which originated in Tunisia in 2011, and the socio-political changes that occurred afterwards. Europeans, and especially the French have long been interested in the region, in relation to the colonial administration notably, since the nineteenth century but the region benefited from new focuses since the beginning of the twenty-first century, especially by the US community of researchers, and even more so since 2011[18]. This explains why studies on the Maghrib in modern and contemporary times are very active nowadays[19]. Even if Maghribī studies traditionally tend to suffer from a certain tropism, because the Maghrib was often regarded as marginal, either an Occidental margin of the Islamic empire whose center would be in Syria or Iraq, or a Southern margin of the Mediterranean, it must be said that new research are currently being carried out on the medieval Maghrib. In that perspective, some projects, notably in Europe, testify to this dynamism. We can mention for example the project *RomanIslam*, led by Stefan Heidemann and his team in Hamburg. This

---

manuscripts. Of the 814 manuscripts, 614 are from Maghreb in Maghribī scripts, that is 67% of the collection so far.

[15] To an overview of the state of digitization of Arabic manuscripts worldwide, we recommend to read Chapter 3 of C. Van Lit's book *Among digitized manuscripts* where he assesses twenty repository including the National Library of Morocco (73-101)

[16] http://bnm.bnrm.ma:86/

[17] http://www.fondation.org.ma/web/accueil

[18] For instance, the creation of the Centre d'Études Maghrébines en Algérie (CEMA) in Oran, 2006, (https://aimsnorthafrica.org/cema/) or more recently, the opening of the Tunisian Office of Center for Middle East Studies from Harvard University in Tunis, 2016 (https://cmestunisia.fas.harvard.edu/).

[19] See notably:
- the theme of the 2022 conference of the Center for Maghrib Studies (http://centermaghribstudies.org/fr/minorities-in-the-maghrib-conference/)
- the conferences organized by The American Institute for Maghrib Studies (https://aimsnorthafrica.org/previous-conference-programs/).
- the Digital Humanities Initiative of the Tunisian Office from Harvard (https://cmestunisia.fas.harvard.edu/cmes-tunisia-digital-humanities-initiative-2020-21).
- the French centers in Maghrib (CJB: https://www.cjb.ma/ ; IRMC: https://www.institutfrancais-tunisie.com/irmc#/) dedicated to the modern and contemporary eras.

project addresses the question of Islamisation processes in the Maghrib and al-Andalus.[20] As for LibMed, a project undertaken by S. Garnier and A. Montel in Lyon, it aims at gaining a better understanding of a region of the Islamic empire which still remains kind of *terra incognita*, namely Libya.[21] These above-mentioned projects do not necessarily showcase manuscripts, but the simple fact that they place the Maghrib at the center of their research allows the manuscripts to be considered somehow.

## 4. The Maghribī Arabic scripts: some historical markers and characteristics

The so-called "Maghribī" Arabic scripts, also called "Western scripts" or "rounded scripts", designate a variety of styles which share common characteristics. Dated back to the tenth century, these scripts were used extensively in the Islamic West – Andalusia and North Africa – as well as in sub-Saharan Africa until the twentieth century. The history and origins of these scripts have aroused significant scientific debate (Bongianino [2017]: vol. I., 9-14; Ben Azzouna [2017]:112-114). At the end of the nineteenth century, O. Houdas was one of the first to study this writing and he made the city of Kairouan, at the heart of present-day Tunisia, the cradle of these scripts (Houdas [1886]). This idea, long accepted, was first questioned by F. Déroche who, at the end of the twentieth century, formulated the hypothesis that the writing of the Egyptian papyri would be at the origin of two types of writing: the so-called Maghribī scripts on the one hand, and the Abbasid bookhands on the other (Déroche [1996]). However, this theory has been quickly dropped in favor of the Andalusian origin of these writings (Bongianino [2017]:12). The most recent works, in particular those of U. Bongianino, have highlighted the different paths (from books to Korans, from Andalusia to the Maghrib) that these scripts took between the tenth and thirteenth centuries (Bongianino [2017]). The rounded characteristics of these spellings could be explained by the instrument used to write, namely calames made from large reeds cut in half lengthwise with one end cut at a point and not at a bevel as in the Orient (Ben Azzouna [2017]: 112).

A family of scripts, the "Western scripts" are characterized by changes over centuries and regions and many paleographical variations that need to be further scrutinized. Ten years ago, M. Jaouhari presented some of his preliminary research on dated manuscripts from Moroccan manuscripts collections, whose he intended to analyze ten centuries of writings (Jaouhari [2013]). In his paper, he focused on manuscripts from the eleventh and twelfth centuries and identified two groups of manuscripts with seven types of writings in the first group and three in the second one. Nonetheless, it seems that he did not pursue his investigation and we do not have at our disposal paleographical studies for pre-modern and modern manuscripts, let alone a complete paleographical understanding of this family of scripts. This lack of substantive work on late scripts therefore leads us to have to take up the traditional and somewhat artificial dichotomies between maghribī *kūfī*, *mabsūṭ*, *thuluth*, *mujahwar* and *musnad* scripts (Afā, al-Maġrāwī [2013]: 57-64). According to this traditional categorization, the maghribī *kūfī* and *mabsūṭ* scripts were the oldest, represented in particular in the writing of the Koranic text in the Muslim West (Afā, al-Maġrāwī [2013]: 57-58). As for the *thuluth* maghribī, it would be an adaptation of the oriental *thuluth* this writing is characterized by a certain stylization of the realizations and is frequently used, in addition to copies of the Koran (Gacek [2009]: 274-275), in the *Dalā'il al-khayrāt*, a famous collection of prophetic litanies of al-Jazūlī (d. 1465).[22] The maghribī *mujahwar* script would have developed at a later time, in the course of the 12th century from the *mabsūṭ*. This type of writing, widely diffused in the extreme Maghrib (*al-Maghrib al-aqṣā*), would not have been limited to literary works but would be used in the writings of everyday life: U. Afā et M. al-Maġrāwī point out that we find this type of script in private letters or official decrees (*al-Ẓahā'ir al-sulṭaniyya*) (Afā, al-Maġrāwī [2013]:62-63). Finally, the maghribī *musnad* script is the name given to a type of writing characterized by its extremely cursivity; this type of writing would be found, in particular, in registers and notarial deeds (Afā, al-Maġrāwī [2013]: 64).

---

[20] *RomanIslam* project: https://www.romanislam.uni-hamburg.de/
[21] LibMed: https://libmed.hypotheses.org/
We can also mention the ERC project led by Corisande Fenwick, whose objective is to study the spread of Islamic ways of life in the Maghrib. These projects gather all the means of historical research, that are: critical scrutiny of literary sources, archeology and to a lesser extent, direct work on manuscripts.

| | |
|---|---|
| *Kūfī* | 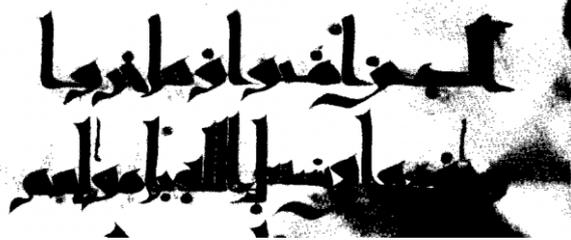 |
| *Mabsūṭ* | 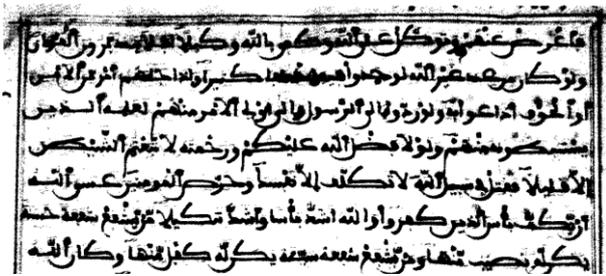 |
| *Thuluth* | 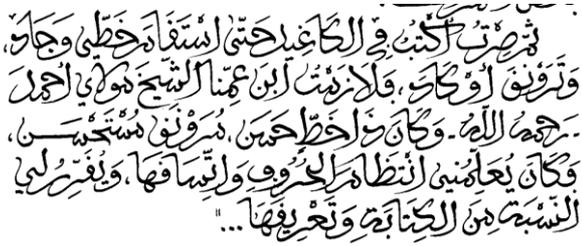 |

---

[22] See below, p. 20

| | |
|---|---|
| *Mujahwar* | 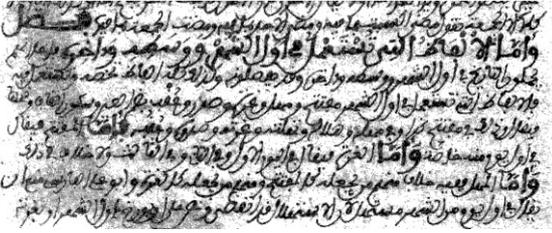 |
| *Musnad* | |

Table 1. Traditional Maghribī script families[23]

What we can assert however is that the Maghribī scripts form a family of round scripts characterized by a certain number of common features, the first of which are very marked curves.[24] These curves are particularly seen in final position for the following letters: *nūn, sīn, shīn, ṣād* and *ḍād*. Their loops will then be very round and pronounced and will, often, tend to overlap on the bottom line (table 2, a.). Another feature of the Maghribī writings is the merging of some letters with others. This is notably the case for the letters *dāl* and *dhal* that merge with the letters *rā* and *zayn* (table 2, b.). It is worth mentioning that these features of the "rounded scripts" can be challenging for handwritten text recognition systems. Other issues that we encountered when working on Arabic Maghribī script manuscripts are not specific to these Maghribī writings. They also pertain to other Oriental languages and handwritten writings. The same word or the same letter can be executed differently in the same manuscript, in the same page and even in the same line (table 2, c.). The spacing between words could also be an important issue: in the example of the table 2, d., where *fī-hi dhālika* is made, the *dhāl* is practically glued to the *fī-hi*. Finally, we also count certain types of abusive and singular ligatures (table 2, e.). In the following pages, we will address other issues and challenges of these manuscripts but with practical and pragmatic approaches, in the process of the HTR workflow.

---

[23] Examples taken from Afā, al-Maġrāwī, *Al-Khaṭṭ al-Maghribī,* 60-61, 100, 104, 109.
[24] See Table 1 in Vidal-Gorène *et al.*, "RASAM," 267. For a detailed presentation of the specificities of these scripts, see: N. Van Den Boogert, "Some notes on Maghribi Script," *Manuscripts of the Middle East* 4 (1989), 30-43; Bongianino, *The Origins and Developments,* 14-16.

| a. | b. | c. | d. | e. |
|---|---|---|---|---|
| 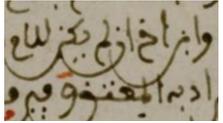 | 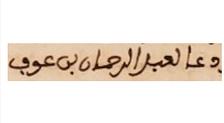 | 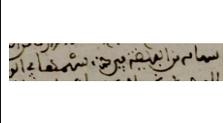 | 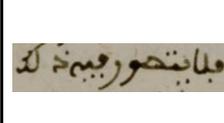 | 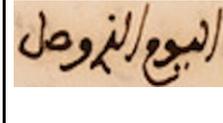 |
| وابن اخ ان لم يكن للام | دعا لعبد الرحمان بن عوف | سهامه من الفريضة جزء سهْمها في في | فلا يتصور فيه ذلك | وصل اليوم الذي |

Table 2. Specificities and issues of the Maghribī scripts for HTR

# II. Versatility of Arabic HTR models for new manuscripts: Methodology and Outputs

In this paper, we aim at assessing the versatility of the HTR model created from the RASAM dataset and to pinpoint under what conditions this model can be sufficiently fine-tuned to cover new Maghribī scripts. In particular, we investigate the volume of data needed to train an efficient specialized model, and its learning velocity. For the layout analysis, we use the models created in September 2021 on RASAM and available on Calfa Vision (Arabic manuscripts project). The scores obtained for the detection of the main text area and the detection of lines are respectively 97.80% (average IoU) and 97.34% (F1-score).

### 1) Looking for the most accurate architecture

The evaluations carried out in the framework of RASAM were undertaken with the architectures proposed by default within the Calfa Vision platform. There are also two other well-established platforms in the scientific world for the transcription of historical manuscripts and documents, namely eScriptorium (developed by INRIA) and Transkribus (now developed by Read-cop). Each of these platforms includes the possibility of training specific HTR architectures, respectively kraken (Romanov, *et al* [2017]), and HTR+ (Michael, *et al* [2018]). In order to select the model that will be used as the basis for its specialization, we trained these three HTRs with RASAM data. In total, five architectures are considered:
- ARA-WB: the word-based architecture proposed by default on Calfa Vision and which was used to conduct the experiments on RASAM. This architecture has notably demonstrated a good capacity to process abbreviated words in medieval manuscript (Camps, Vidal-Gorène, Vernet [2021])
- ARA-CB: the character-based architecture provided by default by kraken;
- ARA-CB+: the character-based architecture provided by default by Transkribus.

These three architectures have already been used to process Arabic scripts and have demonstrated good recognition performance. We add two architectural variants for evaluation:
- ARA-mWB: a word-based architecture proposed by Calfa and used for the processing of Arabic documents where the diacritic signs are seldom and/or inconsistently drawn[25]
- ARA-CB-ext: a variation of the architecture provided by kraken and described within the Lectaurep project.[26]

The selected settings are common to all architectures, thus we are processing images in color, with 120px of picture height, a learning rate of 0.0001, an NFKD unicode normalization without an adaptive learning rate or data augmentation.

### a) Comparison between character-based and word-based architectures

The choice for the core architecture was preceded by a standardization of the training data in order to limit the fragmentation of the recognition and learning process. RASAM contains 7,540 lines of text, and 54 different classes. Some classes being largely under-represented, in particular numbers, vocalization marks (such as *ḍamma* with *tanwīn* with only two occurrences) or *shadda*, we have carried out a standardization of the data to ultimately keep only 38 classes. The deleted characters are not replaced by one or more of the retained classes, which may lead to ambiguity in the text (two different graphic forms may correspond to the same transcription).

---

[25] Architecture currently being developed on the basis of manuscripts written in Kairouan between the ninth and the eleventh centuries. These manuscripts, relatively ancient, are characterized by the lack of or the inconsistency of diacritics, which reveals *problematic* within a character-based approach HTR. Thus, the developed architecture has to rely on a meta-word-based approach.

[26] https://lectaurep.hypotheses.org/488 [11/04/2022]

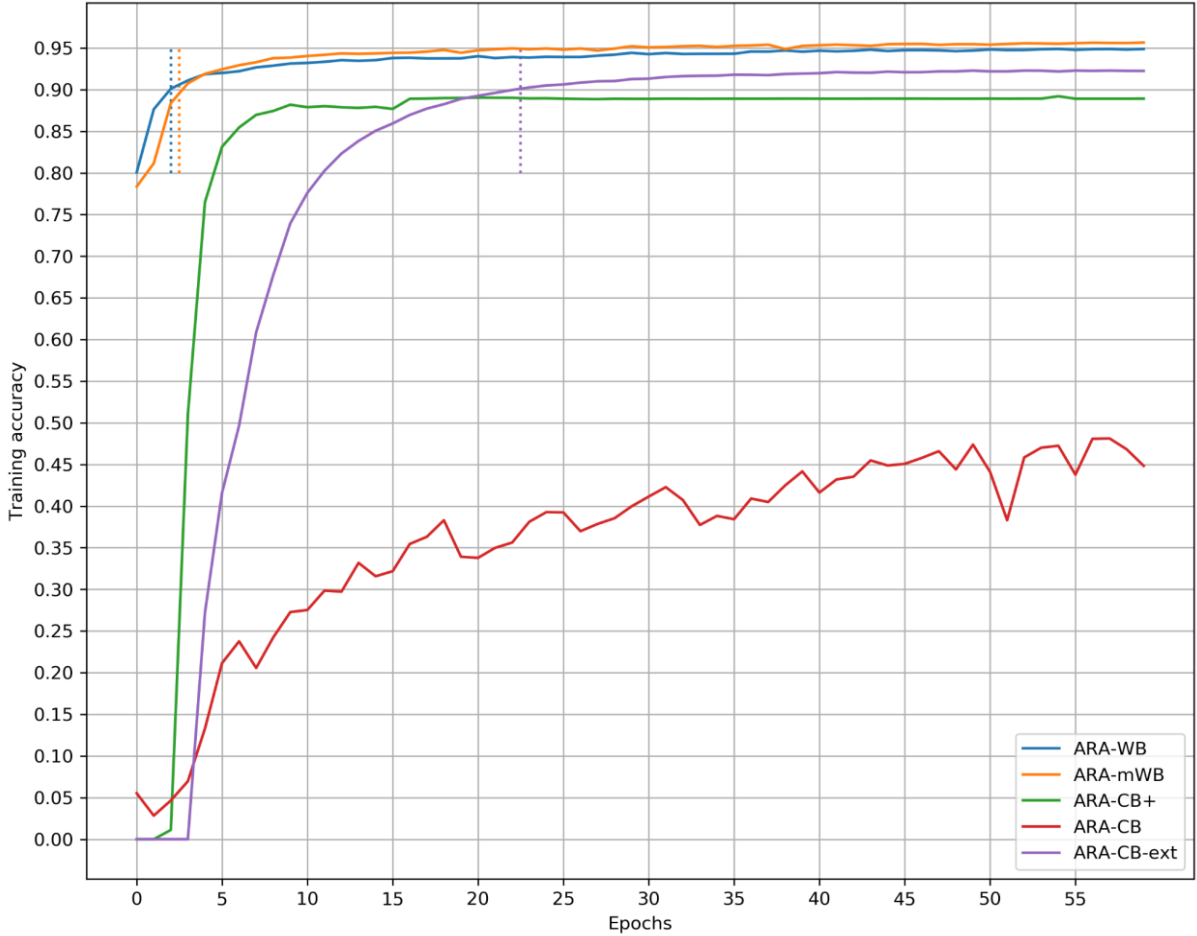

Figure 1. Evolution of training accuracy on RASAM

We observe that word-based architectures converge much faster than character-based architectures, resulting already in an accuracy above 85% after two epochs (one epoch corresponds to the time needed by the neural network to see the whole dataset), whereas the other architectures require between 5 and 12 epochs to reach this level. Throughout the learning process, we notice a very slow progression of the models once the 90% accuracy mark is reached. Nevertheless, ARA-mWB ends up reaching a score of 95.65%, i.e. 3.36% higher than ARA-CB-ext which is the only character-based architecture to exceed 90% accuracy (after more than 20 epochs with 92.29%) and only 0.78% higher than ARA-WB (94.87%).[27]

Although the difference between the final accuracy of the ARA-WB, ARA-mWB, ARA-CB+ and ARA-CB-ext models is not ultimately very significant, the word-based architectures are still more suitable for training models for such cursive scripts. Above all, we observe a learning time that is reduced by a factor of 10, which suggests that the choice of such an architecture should be favored for the rapid availability of efficient models, for a transcription project for instance.

---

[27] This small of a discrepancy in the training data may be due to statistical chance and cannot be deemed significant on its own.

The results of each of these architectures are described in Table 3. The evaluations are conducted on 10 pages from MS.ARA.1977, 7 pages from MS.ARA.609 and 15 pages from MS.ARA.417.

| CER | ARA-mWB | ARA-WB | ARA-CB-ext | ARA-CB+ | ARA-CB |
|---|---|---|---|---|---|
| All | 3,23 | 4,03 | 7,39 | 9,51 | 53,69 |
| MS.ARA.1977 | 3,08 | 3,42 | 6,57 | 9,87 | 58,01 |
| MS.ARA.417 | 2,81 | 2,77 | 6,01 | 10,70 | 23,34 |
| MS.ARA.609 | 3,88 | 5,40 | 11,71 | 15,13 | 57,59 |

Table 3. Final CER on the whole standardized RASAM testing set

We do not notice a significant difference between ARA-WB and ARA-mWB. The overall CER remains in the same proportions. MS.ARA.609 alone benefits from the ARA-mWB model, with a reduction in CER of 1.52 points. Although the results obtained by character-based architectures are quite usable in a transcription project, in particular for MS.ARA.1977 and MS.ARA.417 with the ARA-CB-ext architecture, the CER of these approaches remains at least twice as high as for word-based architectures. The MS.ARA.609 manuscript shows a higher complexity (see below), that the ARA-mWB architecture alone is able to cope with here, as can be seen in Figure 2 (see below).[28]

b) **Challenges and observations**

The RASAM manuscripts have difficulties on several levels that the overall CER does not reflect. Notably, this being the case for marginalia and catchwords, often slanted, with a more cursive meticulous script. The error analysis of the first model created from RASAM highlights the difficulty to recognize texts in color and semantic and/or decorative symbols. This phenomenon is displayed on Figure 3 and poses a particular challenge for the creation of HTR models. The word-based models prove to be substantially more stable than the character-based ones.

---

[28] MS.ARA.609 shows a higher complexity due to multiple factors. Among these, a tighter and more linked script, for which the notion of characters is more difficult to define, as well as shorter writing lines due to the inclusion of tables and also many interlinear overlaps. Colors scripts (red notably) also need to be mentioned.

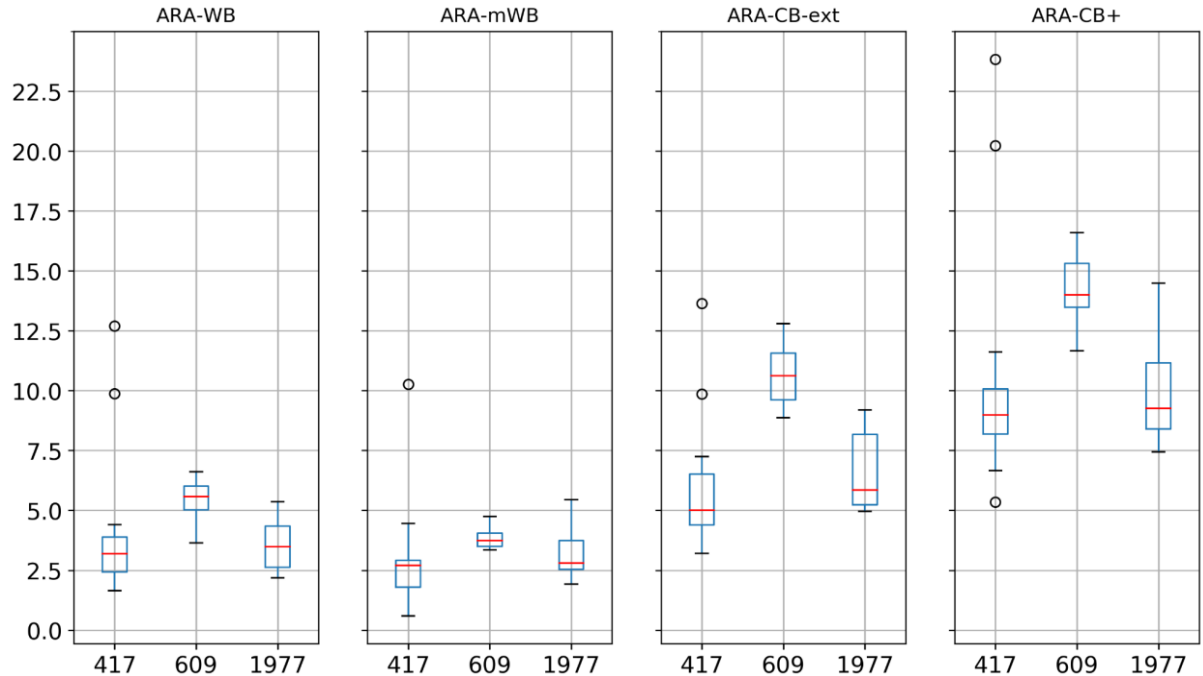

Figure 2. Error distribution for each architecture on each manuscript from the testing set of RASAM[29]

We observe on Figure 2 that the CER median is 3.625% for ARA-WB, 2.915% for ARA-mWB, 6.21% for ARA-CB-ext and 9.875% for ARA-CB+, it provides further evidence of the significant benefit of word-based approaches for these scripts. The boxplots of architectures ARA-WB and ARA-mWB are substantially similar, although we can notice improved performance for the outcomes of MS.ARA.609. The character-based architectures provide usable results nonetheless, with a CER as low as 3.2% for some images of MS.ARA.417 and as low as 4.96% of MS.ARA.1977 for ARA-CB-ext. Finally, a quarter of all images achieve a CER below 2.6% with a word-based architecture, with a minimum of 0.6% reached by ARA-mWB on page 42 of MS.ARA.417.

Nevertheless, in the MS.ARA.417 manuscript, the pages 5 and 9 appear to be problematic for all architectures, with a respective CER of 12.70% and 9.87% for ARA-WB, 13.64% and 9.86% for ARA-CB-ext, and finally, 23.84% and 20.23% for ARA-CB+. The ARA-mWB architecture alone is successful in recognizing adequately page 9 (CER of 4.46%) but fails with page 5 with a CER of 10.27% (see Figure 3). Although the errors on page 5 can be explained easily due to a marginalia very differently handwritten, lines completely vocalized and semantic symbols in red that serve to separate sentences or sections, page 9 does not seem to display any particular difficulty. The page, however, holds semantic or decorative symbols in red, that seem to be the source of the difficulties encountered by the character-based approaches.

---

[29] The boxplots allow, for a given data set, to depict its median, its quartiles and its extremum values. From bottom to top, we can read the minimum value (the smallest CER achieved), the first quartile, the median (in red, half the evaluated lines obtain a CER lower than this value), the third quartile, and the maximum value (the highest CER obtained). The outlier results are depicted as outside points. A constricted boxplot indicates the model's great stability and the consistency of results, contrary to an expanded boxplot.

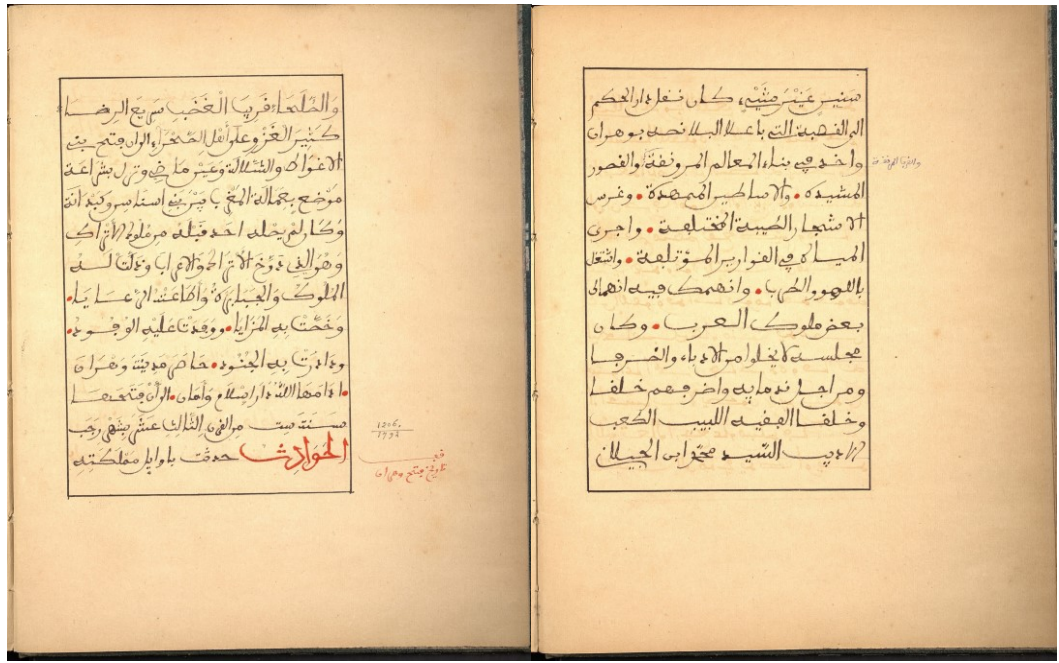

Figure 3. MS.ARA.417, p. 5 (left) and MS.ARA.417, p. 9 (right)

| MS.ARA.417, p.9, ground truth | ARA-mWB prediction<br>CER: 3,49% | ARA-WB prediction<br>CER: 7,69% |
|---|---|---|
| المشيده والاساطين الممهدة وغرس الطيبة المختلفة واجرى الاشجار المياه في القوارير المؤتلفة واشتغل باللهو والطرب وانهمك فيه انهماك بعض ملوك العرب وكان | والاساطين الممهدة وغرس المشيد واجرى المتلفة الىية الا شجار واشتغل المياه في القوارير الموتلفة باللهو والطرب وانهمك فيه انهماك بعض ملوك العرب وكان | الممهدة وغرس والاساطين المشيد واجرى المتلفة الطية الاشجار واشتغل المياه في القوارير الموتلفة انهماك وانهمك قيه باللو هووالطرب وكان العربا هلوك فعض |
|  | ARA-CB-ext prediction<br>CER: 11,88% | ARA-CB+ prediction<br>CER: 32,86% |
|  | الممهدة وغرس المشيده والاساطينر الطيبة المختلفة واجرن الا شجار واشقل الموتلعة الغوارير المياهفي انهمال فين والطربوانهمك باللو وكلان ملوك العزبا فعض | وغرس اللممهدكة والاسا ط الحشيهه واجرى الطيبة المغنلعت الا شحار واشنذل البمقتلععة في الغواري الباه انهماك فىى يك باللمموو الطرربوامهدي هلو سك العربيد كن فيبعض |

Table 4. Comparison of the prediction of each model on page 9 of BULAC MS.ARA.417

Table 4 shows that the ARA-CB-ext and ARA-CB+ architectures encounter difficulties to predict spaces. In both cases, the models struggle to separate words. Moreover, semantic and decorative symbols add a real challenge for these architectures, with regards to the segmentation. The example of the fourth line is significant: these two architectures link والطرب to وانهمك where a space was expected. Conversely, a good word separation is noticed with the ARA-WB and ARA-mWB architectures which, each, however record a separation error: باللهو والطرب for ARA-WB and الاشجار for ARA-mWB.

One might wonder to what extent the fact that ARA-CB-ext and ARA-CB+ are architectures that are not specialized for Arabic Maghribī scripts may explain some of the prediction errors that we observed on some characters. These architectures seem to struggle to predict letters that have, in the Arabic Maghribī scripts, a diacritic dot placed above the character like the letters ق and غ, or خ. Thus, we see that القوارير is systematically predicted الغوارير, with ARA-CB-ext et ARA-CB+. This is also the case for letters where the diacritic dot is placed below the character, instead of above in Oriental scripts, like the letter ب. In the word المؤتلفة, the letter ب is predicted as a ع with ARA-CB-ext and ARA-CB+. At this stage, the sample is not sufficient to reach any final conclusions, but the observation is worth mentioning.

For the other manuscripts, we observe a greater proximity between the results obtained by ARA-WB and ARA-mWB. If we take two classically less successful examples with the RASAM model, we notice a slight CER improvement and better word separation. All the same, both models still encounter difficulties with the recognition of scripts in red ink, especially on MS.ARA.609.

| 1977 p. 16, l. 22 | 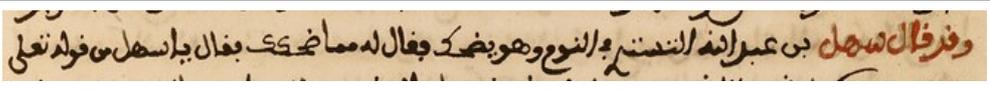 |
|---|---|
| GT | وقد قال سهل بن عبد الله التستري في النوم وهو يضحك فقال له مما ضحكك فقال يا سهل من قوله تعلى |
| ARA-WB CER : 3,29% | وقدقال سهل بن عبد الله التستر في النوم وهو يضحك فقال له مما ضحكك فقال ياسهل من قوله تعلى |
| ARA-mWB CER : 1,09% | وقد قال سهل بن عبد الله التستر في النوم وهو يضحك فقال له مما ضحكك فقال يا سهل من قوله تعلى |

| 609 p. 24, l. 3 | 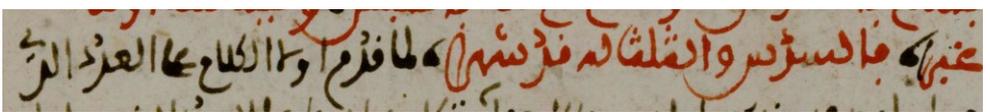 |
|---|---|
| GT | غبرا فالسدس والثلث له قد قدم شهرا او ولا الكلام على العدد الذي |
| ARA-WB CER : 9,09% | غبره فالسدس والثلث له قد شهراه لما قدم اولا الكلام على العدد ال |
| ARA-mWB CER : 7,57% | غبرا فالسدس والثلث له قد شهرا لما قدم اولا الكلام على العدد ا |

Table 5. Comparison of predictions between word-based and meta word-based approaches on MS.ARA.609 and MS.ARA.1977 manuscripts

The fact that these scripts are rather narrow and condensed, and sometimes give the impression that there is no separation between words, do not seem to be a challenge for the different architectures. Therefore, word separation is accurate with 95.20% on average for the word-based architecture, whereas it decreases to 91.16% for the character-based architecture. Prediction errors focus on specific points like ligatures. It is evident in the case of the ligature between the letters *rāʾ* and *yāʾ* in final position – see التستري in MS.ARA.1977 (table 5) – or when the letter *yāʾ* subscribed to the baseline – see الذي in MS.ARA.609, at the end of the line (table 5). Contrary to ARA-WB, ARA-mWB achieves better results regarding words separation: ARA-WB did not predict the space between قد and قال, or between يا and سهل in MS.ARA.1977, p. 16, l. 22 whereas ARA-mWB did. On top of that, semantic and/or decorative symbols constitute an issue for ARA-WB, which predicts these symbols like the letter ه (see MS.ARA.609, table 5), the meta-word based approach offers better results and does not predict them as expected.

The ARA-mWB architecture, developed by Calfa, appears less sensitive to recurring noise present in the data of RASAM. Therefore, this architecture seems to be the most relevant one to deal with the writings under consideration that are Arabic Maghribī scripts. Even though the benefit in terms of CER is limited to 1%, this architecture shows a higher adaptability.[30] The character-based architectures, although capable of providing useful results, do not succeed in covering the diversity of RASAM with the same amount of data, and are less consistent. Figure 4 shows the overall error distribution obtained with the ARA-mWB architecture on the standardized RASAM dataset. Generally speaking, character prediction is effective but the confusion matrix enables to pinpoint recurring errors. Overall, they are directly linked to the *hamza* and its realizations: we notice that the ARA-mWB architecture encounters some difficulties to distinguish the *alif* from the *hamza* and usually mistakes ا, أ and إ. To a lesser extend, this architecture can erroneously predict the character ؤ, that will be predicted و.

These results – taking into account the few limitations underlined above – comfort us to choose the ARA-mWB architecture as our entry-level model. Our experiments will be based on it and we will fine-tune it on the new handwritings considered.

---

[30] At this stage, it is not possible to conclude that the new architecture resolves all the issues raised by RASAM. The CER of 3.23% achieved is indeed lower than the CER achieved in the RASAM article, yet these findings need to be confirmed by further experiments.

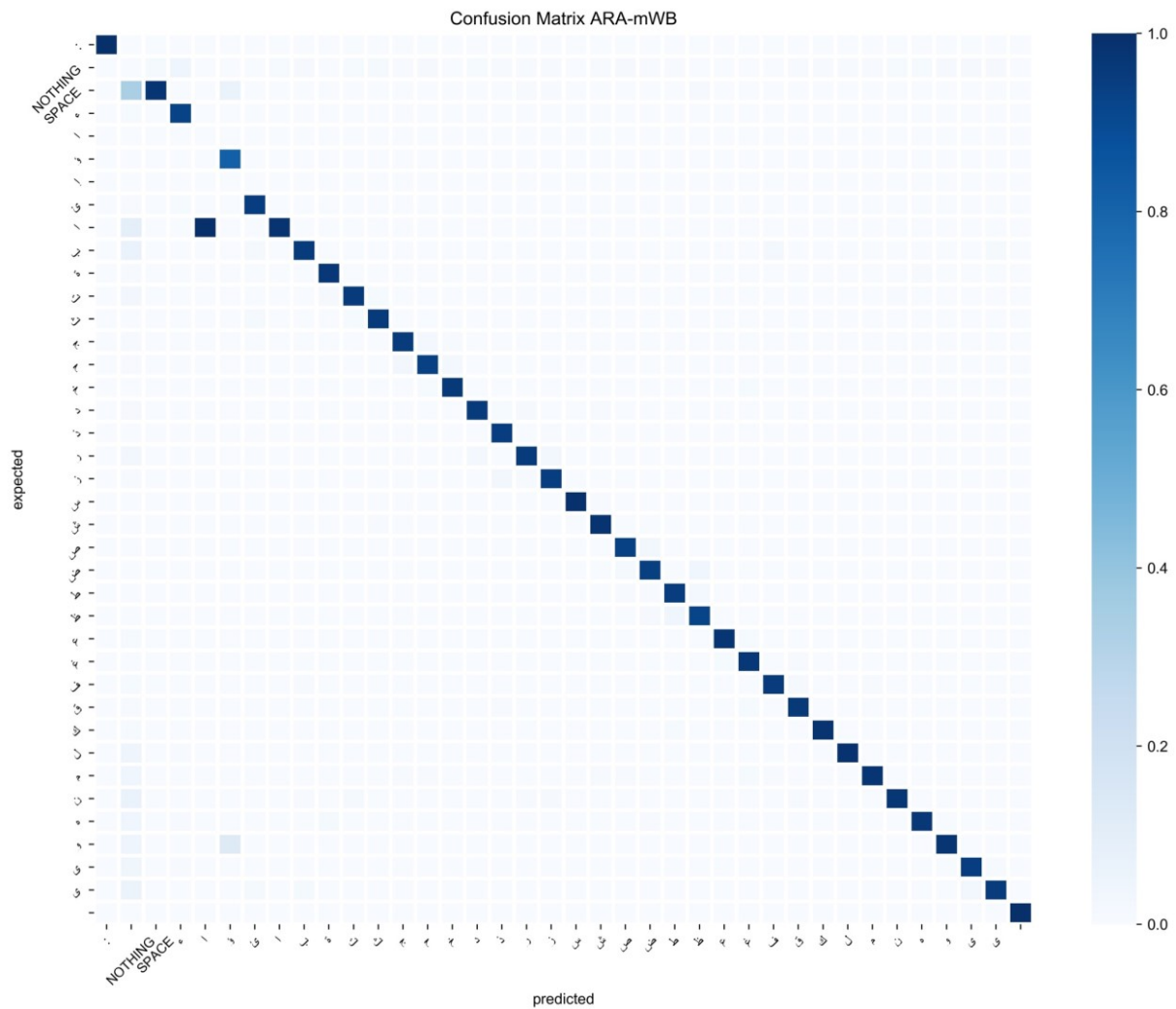

Figure 4. Confusion matrix for the ARA-mWB model on the standardized RASAM dataset

### c) Intermediate conclusion

Once standardized, as described above and with the choice of the ARA-mWB architecture, we obtain a mean CER of 3.23%, that is an accuracy of more than 95%, thus favoring this architecture in our following experimentation. We also discussed challenges that participate in explaining remaining mispredictions. At this point, it is worth summarizing what we assume are the reasons why we can not exceed 3.23%:

- Colored inks are less recognized than black ink. So far, we mainly notice red ink problems, but the RASAM dataset also comprises some green ink that show similar problems.
- Signs, whether decorative or semantic (used to delineate or to separate verses of poetry for example), might introduce noise for the algorithm. At a larger scale, all extra signs, vocalizations included, can introduce difficulties.
- Spaces between words in a context of *scripto continua* remain sometimes a challenge, even though the meta-words based approach rectifies somehow.
- Margins notes, catchwords are usually less recognized

- Specific ligatures, especially in final position in the words cause mistakes

Do these remaining challenges can be resolved with more and diversified data, or are they a substantive issue in terms of HTR of Arabic Maghribī script manuscripts?

### 2) Fine-tuning the model on new scripts

#### a) Corpus for research: Selection and characteristics

The versatility of the RASAM model has been tested on six manuscripts from BULAC's digital collections, which were selected according to three criteria.[31] The first criterion was the theme of these manuscripts. In order to diversify the initial RASAM dataset which includes a law book (MS.ARA.609) and two historical books (MS.ARA.417 and MS.ARA.1977), we chose manuscripts dealing with different topics. Consequently, our aim was to evaluate the versatility of the model created on a lexicon restricted to historical and legal themes on a plurality of other themes. It enables us to assess the possibilities of prediction from existing models on manuscripts with potentially unknown vocabulary due to their topic (lexicons related to grammar or erotology for example). Among the six manuscripts selected, only two of them share themes with the RASAM dataset, MS.ARA.1944 and MS.ARA.1957, offering comparison prospects.

Therefore, we tested RASAM model on the following manuscripts:

1. The **MS.ARA.1922** comprises a treatise on erotology entitled *al-Rawḍ al-ʿāṭir fī nuzhat al-khāṭir* composed by Cheikh Muḥammad al-Nafzāwī (d. around 1434). The manuscript (235 x 180 mm), copied in 1892, consists of 92 folios, each page having 13 lines.[32]

2. The **MS.ARA.1925** is a collection of examples of correspondence between dignitaries or between individuals and dignitaries (kings, scholars, saints) in prose and verse. The manuscript (180 x 120 mm), composed at an unknown date, has 10 folios, each page has 11 lines.[33]

3. The **MS.ARA.1926** records the famous collection of prophetic litanies entitled *Dalāʾil al-khayrāt* of al-Jazūlī (d. 1465) (Abid [2017]). Copied in 1891, this 152-

---

[31] Manuscripts available on the BINA website (BULAC digital library, inaugurated in January 2019: https://bina.bulac.fr/). BINA currently provides 251 Arabic manuscripts. The six manuscripts selected for this research come from the collection of Paul Geuthner and Warburga Seidl, established as part of the activities of the Orientalist Librarian Paul Geuthner founded in 1901. This collection was acquired by the BULAC in November 2016 from the Galerie Laure Soustiel. It holds 88 manuscripts, including 38 in Maghribī script.
[32] Link BINA: https://bina.bulac.fr/ARA/MS.ARA.1922
[33] Link BINA: https://bina.bulac.fr/ARA/MS.ARA.1925

folio manuscript (200 x 175 mm) includes illuminations and decorations in yellow, blue, red and green ink. The text is spread over 11 lines for each page.[34]

4. The **MS.ARA.1929** comprises Ibn Hishām's (d. 1360) treatise on grammar entitled *Sharḥ qaṭr al-nadā wa-ball al-ṣadā*.[35] On watermarked paper, this incomplete manuscrit (235 x 175 mm) of 26 folios produces the text in two colors: the text of the *Qaṭr al-nadā wa-ball al-ṣadā* is in red, the author's comments (*Sharḥ*) are in black. The text runs over 29 lines.[36]

5. The **MS.ARA.1944** consists of a history of Tunisia under the Almohad and Hafsidd dynasties (*Tārīkh al-dawlatayn al-muwaḥḥidyya wa-l-ḥafṣiyya*) of al-Zarkashī (d. 1477 or 1489) (Garnier [2022]: 85-86). The manuscript (225 x 165 mm) consists of 30 folios, each page having 20 lines.[37]

6. Finally, the **MS.ARA.1957** is a compendium of Mālikī law (*al-Mukhtaṣar*) by the Egyptian jurist Khalīl b. Isḥāq al-Jundī (d. 1365).[38] Composed on paper, this manuscript (150 x 160 mm) consists of several notebooks sewn together. The text runs 15 lines per page.[39]

| 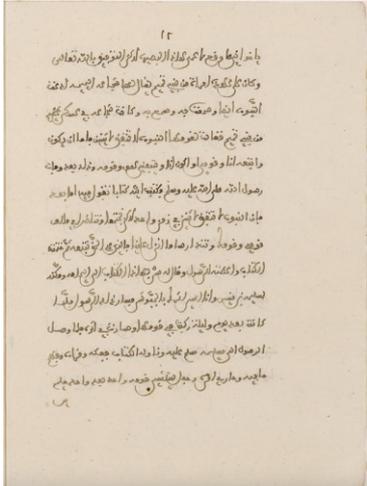 | 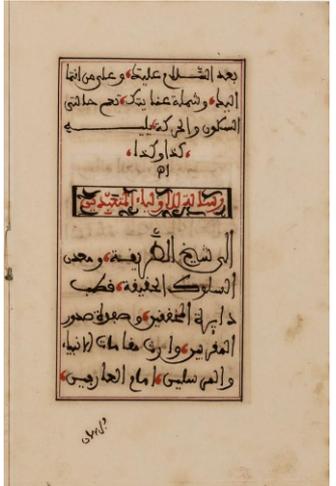 | 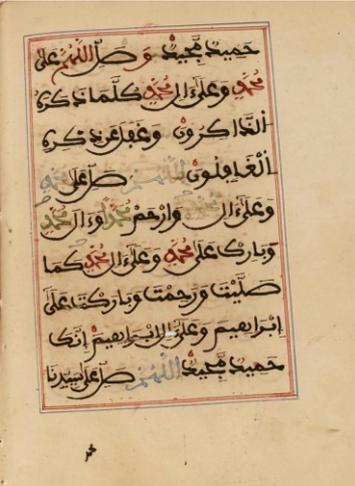 |
|---|---|---|
| MS.ARA.1922, p. 14 | MS.ARA.1925, p. 8 | MS.ARA.1926, p. 142 |

---

[34] Link BINA: https://bina.bulac.fr/ARA/MS.ARA.1926
[35] On this scholar, see: H. Fleisch, "Ibn Hishām", *EI².*
[36] Link BINA: https://bina.bulac.fr/ARA/MS.ARA.1929
[37] Link BINA: https://bina.bulac.fr/ARA/MS.ARA.1944
[38] On this scholar, see: M. Ben Cheneb, "Khalīl b. Isḥāk", *EI².*
[39] Link BINA: https://bina.bulac.fr/ARA/MS.ARA.1957

| | | |
|---|---|---|
| MS.ARA.1929, p. 38 | MS.ARA.1944, p. 14 | MS.ARA.1957, p. 224 |

Table 6. Examples of the six selected manuscripts

In addition to themes, the second selection criterion was the variety of layouts and writings. The selected manuscripts offer varied layouts. Some are very airy with less than ten lines per page and a few words per line (MS.ARA.1926) while others are much denser with around thirty lines per page and many words per line (MS.ARA.1929). Some manuscripts do not show any marginal note and present a text centered in the middle of the page (MS.ARA.1925), whereas others see a multiplication of comments in the margins or even between the lines (MS.ARA.1957).[40]

The last criterion pertains to the Maghribī scripts *per se*, in the sense of how this script was written by the scribe. We intend to evaluate the versatility of the RASAM model on a plurality of writings all encompassed in the "Maghribī" family of scripts. As a result, we focused on different hands and different styles: the MS.ARA.1925 presents a very calligraphic writing like the MS.ARA.1926, whereas the MS.ARA.1957 presents a very book-hand handwriting and the MS.ARA.1929 a more informal one.

---

[40] If all the manuscripts in the RASAM dataset offered marginal notes, none however showed comments developing between the lines.

| | RASAM Manuscripts | | | Tests manuscripts | | | | | |
|---|---|---|---|---|---|---|---|---|---|
| | MS.ARA. 1977 | MS.ARA. 609 | MS.ARA. 417 | MS.ARA. 1922 | MS.ARA. 1925 | MS.ARA. 1926 | MS.ARA. 1929 | MS.ARA. 1944 | MS.ARA. 1957 |
| في | | | | | | | | | |
| هذا هذه | | | | | | | | | |
| الذي | | | | | | | | | |
| على | | | | | | | | | |

Table 7. Different realizations of several words between the RASAM dataset and the six manuscripts of the evaluation

An efficient approach, in order to quickly create new HTR specialized models, consists in fine tuning an existing model. In other words, we start from a pre-trained model that is specialized on a new task. This approach is classic and is implemented in most computer-assisted transcription projects (Vidal-Gorène, *et al* [2021a]). While it is still necessary to provide the model with pages that have already been transcribed and annotated, the volume required is much lower than that needed to train a model from scratch. The aim of this section is therefore to assess the data input required to process a representative variety of Maghrebi Arabic scripts, based on RASAM. We take on purpose the case of the processing of a very poorly endowed graph for which it is not possible to provide a large training database.

**b) Selection of the transcription device and levels of annotations**

The testing was undertaken with the Calfa Vision platform, which, since the hackathon that resulted in RASAM, has integrated pre-trained models for Maghribī Arabic scripts (layout analysis models, and the HTR ARA-WB model described above). These models are specialized as the transcription project progresses.

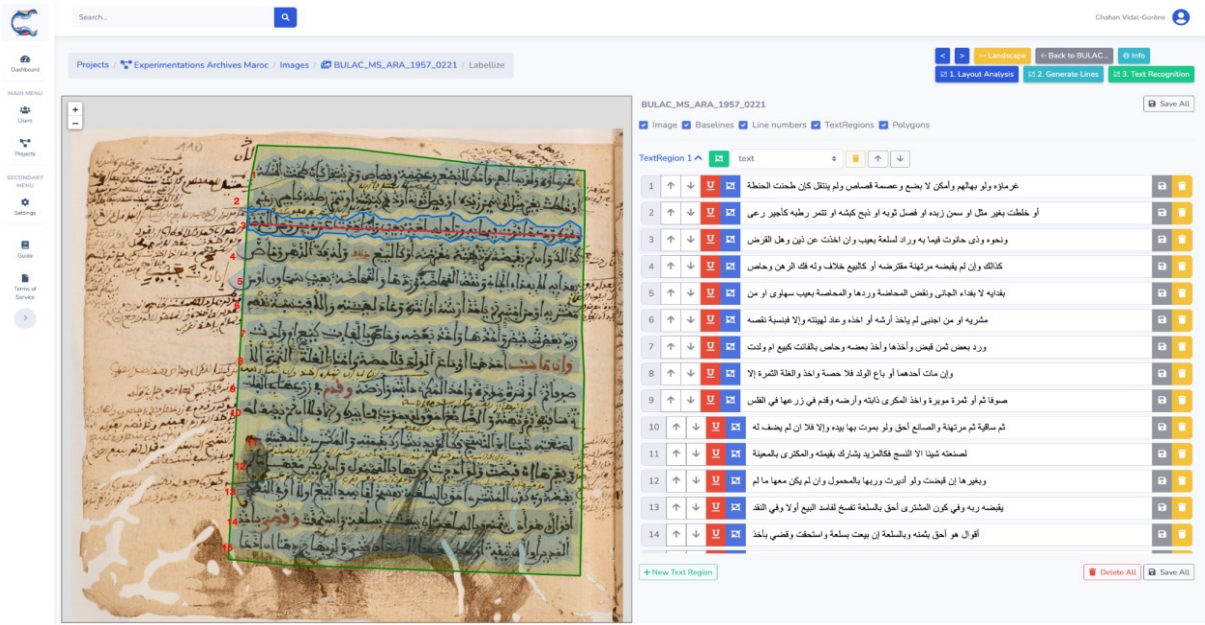

Figure 5. Calfa Vision – platform for computer-assisted handwritten documents transcription

In the framework of our multiple tests, we concentrate on the main text and catchwords. As seen on figure 5, we do not take into consideration margins notes and interlinear additions. In order to keep complete monitoring on the HTR workflow and for the purpose of maximizing the recognition results, we have chosen a three-level annotation:

- the **text region**, which matches the area of interest where the text to be detected is located (in green in figure 5);
    - the **text line**, marked with a baseline (in red) and extracted with a surrounding polygon (in blue);[41]
        - the **transcription** itself, following the specifications defined for RASAM (ref RASAM) with the same standardization applied in this article.

Therefore, we choose a highly supervised annotation where each object is precisely identified in the image. While there are other possible approaches, including on Calfa Vision, such as transcribing a page without the precise identification of lines, this approach greatly reduces the learning problem for the models, limits the amount of data needed, and results in more quickly usable HTR models.

The annotations are first obtained automatically thanks to the preexisting models on Calfa Vision, then the results are manually verified. The models may provide more complete annotations than required for the purposes of the project (identification of margins notes for instance). The manual checking and proofreading of the predictions and the choice to limit to certain text regions (like the main text or the catchwords) result in a progressive adaptation of the platform to the project needs. It translates into predictions gradually closer to the annotation requirements defined at first.

---

[41] Precisely, the baseline is the fictional line upon which the text line lies. The identification of baselines, as a prerequisite to the line extraction, enables to grasp the curvature of the text.

**c) Models fine-tuning and results**

It is recalled that the base HTR model (ARA-mWB$_{base}$) has been trained with the manuscripts MS.ARA.417, MS.ARA.609 and MS.ARA.1977. Although it has demonstrated good learning ability with a final CER of 3.23% (see *supra*), it is still very specialized for these manuscripts. The immediate implementation of ARA-mWB$_{base}$ on the manuscripts selected yields very heterogeneous results, and is a demonstration of the versatility of the model on "out-of-domain" data:
- CER of ARA-mWB$_{base}$ on MS.ARA.1922: 29.54%
- CER of ARA-mWB$_{base}$ on MS.ARA.1925: 15.00%
- CER of ARA-mWB$_{base}$ on MS.ARA.1926: 25.62%
- CER of ARA-mWB$_{base}$ on MS.ARA.1929: 11.97%
- CER of ARA-mWB$_{base}$ on MS.ARA.1944: 14.85%
- CER of ARA-mWB$_{base}$ on MS.ARA.1957: 30.46%

Aside from the MS.ARA.1929 manuscript, for which the 11.97% rate is sufficient enough to achieve usable text recognition, the error rate remains very high for the others manuscripts, up to one erroneous character out of three for MS.ARA.1957. RASAM default settings do not allow to cover a large variety of Arabic scripts. One effective strategy consists in fine-tuning the model to the needs of the new manuscript, increasingly specializing it to the specificities of the scan, the hand of the copyist or the state of preservation of the document. This iterative specialization strategy was already adopted and successfully implemented for the hackathon, resulting in an improvement of the CER after 20 pages and in a 42% reduction of the proofreading time throughout the hackathon (Vidal-Gorène, *et al* [2021b]: 277).

We undertake this iterative approach once more, while significantly decreasing the amount of data processed. The models have been fine-tuned every 20 lines (that corresponds to an average manuscript page length) in order to benefit more quickly from the fine-tuning and to restrict the amount of data needed, which is critical when processing poorly-endowed scripts.[42] The results are displayed on Figure 6. We are achieving one model for each manuscript and for each data batch without mixing data. Each new model is evaluated on a fixed training set, representative of the production from the targeted manuscript, and holding on average 88 lines manually transcribed.

---

[42] By contrast, the fine-tuning approach for Latin scripts requires between 150 and 2,500 lines, with base models trained on even larger volumes. With a ratio of 1 to 10 for the speed of specialization, we achieve a higher CER admittedly, but also a greater complexity in the recognition.
C. Reul, S. Tomasek, F. Langhanki, and U. Springmann. "Open Source Handwritten Text Recognition on Medieval Manuscripts using Mixed Models and Document-Specific Finetuning". arXiv:2201.07661 [cs], 19 janvier 2022. http://arxiv.org/abs/2201.07661.

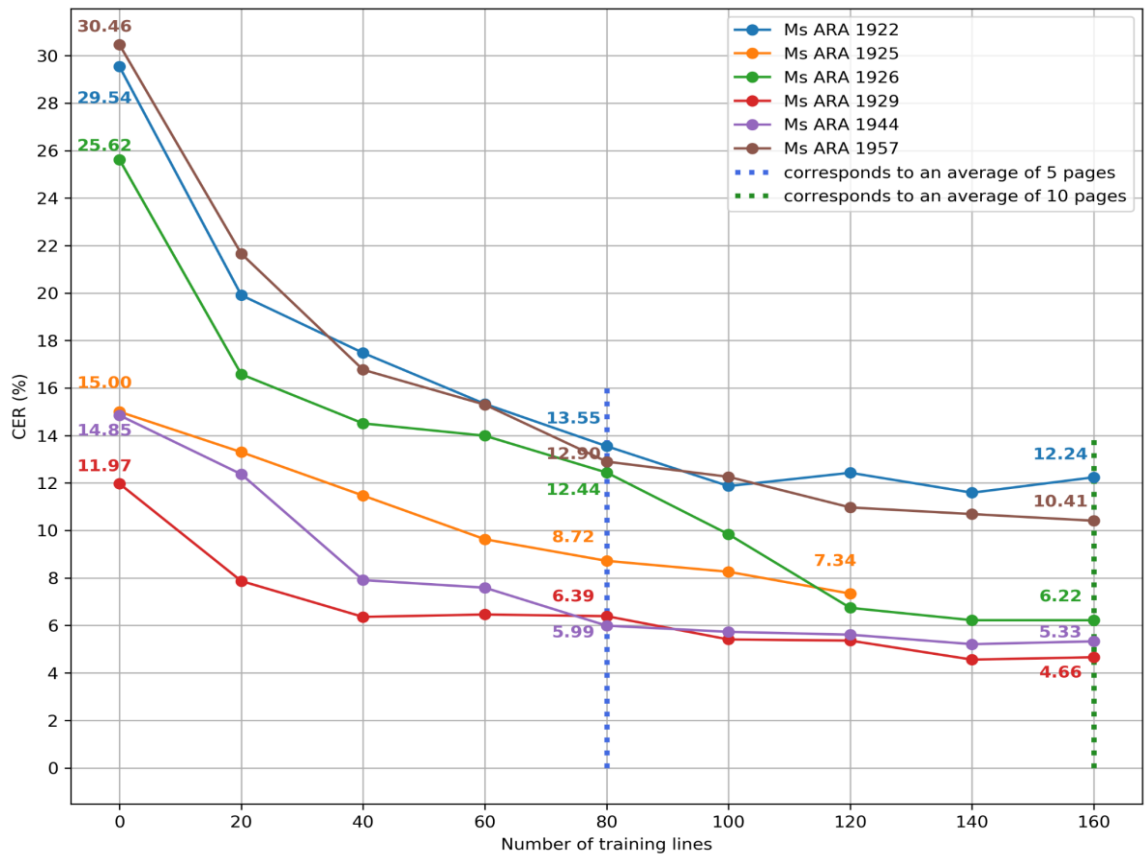

Figure 6. CER Evolution depending on the quantity of transcribed lines

The assessment demonstrates a rapid specialization of the models, with ARA-mWB$_{base}$ achieving an initial average CER of 21.24% whereas ARA-mWB$_{80}$ reaches a CER of 9.99%, i.e. an error rate divided by 2.1 with only 5 pages transcribed, with scores below 10% for MS.ARA.1925 (8.72%), MS.ARA.1929 (6.39%) and MS.ARA.1944 (5.99%).

The MS.ARA.1926 manuscript, which has only 9 lines per page and a particularly large and calligraphic handwriting, achieves scores similar to the other manuscripts with twice the amount of data. On average, the final CER obtained is 7.7% for an average of 10 transcribed images (i.e. on average 160 lines).

The CER of some manuscripts may peak after 100 transcribed lines, or even rise slightly, which can be explained by the chosen approach, i.e. the gradual transcription of pages within the framework of a transcription project, leading to a non-homogeneous distribution of data in the datasets considered iteratively (e.g. presence of numerous titles in the new transcribed page). We observe this phenomenon in particular in table 8, line 2.

| MS.ARA.1926 | Prediction ARA-mWB$_{base}$ | Prediction ARA-mWB$_{1926-80}$ | Prediction ARA-mWB$_{1926-160}$ |
|---|---|---|---|
| 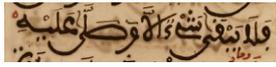 | CER: 15.38%<br>فلا يتقتى شتالا وصلى عل | CER: 11.53%<br>فلا يبقى شى ءالاوصلى عل | CER: 3.84%<br>فلا يبقى شى ء الا وصلى عل |
| 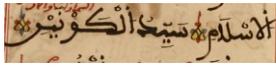 | CER: 21.05%<br>الاسلام ه سيت الكون | CER: 10,52%<br>الاسلام سيد الكون | CER: 5.26%<br>الاسلام  سيد الكون |
| 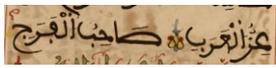 | CER: 0%<br>عز العرب صاحب الفرج | CER: 10.52%<br>عن الفرب صاحب الفرج | CER: 15.78%<br>عن القرب  صاحب الفرج |
| 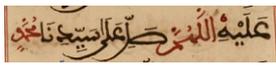 | CER: 28.57%<br>عليه اللمث ل علا سيدنا عذن | CER: 28.57%<br>عليه اللنر ل على سيد نا عى | CER: 0%<br>عليه اللهم صل على سيدنا محمد |
| 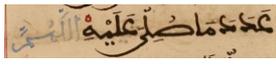 | CER: 19.04%<br>عدد ما صل عليه ال | CER: 19.04%<br>عددما صلى عليه ال | CER: 4.76%<br>عددما صلى عليه اللهم |
| 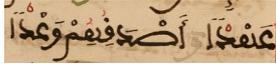 | CER: 35.29%<br>عق ا اصد قهن وعدا | CER: 11.76%<br>عهدا اصدقهم وعدا | CER: 5.88%<br>عهدا أصدقهم وعدا |
| 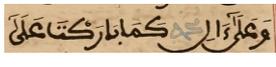 | CER: 18.51%<br>وعلى ءال   كما باركت على | CER: 18.51%<br>وعلى ءال   كما باركت على | CER: 0%<br>وعلى ءال محمد كما باركت على |

Table 8. Evolution of the text recognition throughout the annotation project

In general, Table 8 illustrates the evolution and improvement of the model over time, batch after batch. Quite logically, we see that fine-tuning leads to a better prediction of the characteristics specific to this manuscript that we do not find in the base RASAM model – the use of different colors in particular. The prediction of these colored words testifies to the improvement of the model. In fact, the base architecture had minimal training to predict words or characters in color which, in the MS.ARA.1926, are present at almost every line. Hence, with the ARA-mWB$_{base}$ architecture, the model is falling short of predicting correctly the terms اللهم on lines 5 and 6, and محمد on lines 5 and 7: the specialization after fine-tuning on 160 lines (about ten pages) leads to a good recognition of these terms. If, in general, a meta-word-based approach allows for a better recognition of colored writings, it is worth mentioning however that too light colored writings can still prove to be problematic, even after specialization: in none of the batches, for example, the و realized in yellow at line 6 is predicted.

The results of Table 8 confirm, moreover, our observations on the decoratives and semantics signs: although, with the basic model, these symbols can be understood as the realization of the letter ه (see in particular the prediction الاسلام ه for line 2 with ARA-mWB$_{base}$), fine-tuning enables the model to understand that these signs should not be predicted: then with ARA-mWB$_{1926-80}$ and ARA-mWB$_{1926-160}$ there is no longer any prediction for this symbol — although with ARA-mWB$_{1926-160}$ the model inserts an additional space.[43] The CER which, for a given page, evaluates the difference between the predicted text and the expected result (ground truth), detailing the minimum number of character insertions, substitutions, and deletions to get from one to the other, consequently, is worse – it goes from 0% to 5.26% even though there is, in this particular case, no prediction error.

---

[43] The insertion of an additional space, instead of the semantic or decorative symbols is also visible in ARA-mWB$_{1926-160}$ line 3, where the model inserts two spaces between القرب and صاحب.

We find some small sporadic errors in the prediction of spaces. If, in general, a meta-word-based approach offers a better word-separation than a character-based approach, the fact remains that the model can create spaces where none are expected or, on the contrary, suppress a space where it is expected. This is the case, in particular, in the realization of the word شيء in line 1 where the *hamza* is separated from the word or, again, for عدد ما in line 5 where a space was expected between the two words. In this second example, apart from this missing space, the prediction is correct: again, the CER is not fully representative of the result. In addition, and as minor errors, the noise that may occasionally appear on the manuscript can lead to a prediction error: the dot that appears above the *lām* of العرب in line 2 — which, presumably, is present here to mark the absence of a vowel on this letter — is associated with the following letter: with ARA-mWB$_{1926-80}$, the model predicts *'ayn* as a *fā* and then, in ARA-mWB$_{1926-160}$, as a *qāf*, whereas with ARA-mWB$_{base}$ — without specialization — the letter had been correctly predicted.

This prediction error, as well as the one found in the same line in the confusion between عز and عن, shows however the limits of model specialization. The model had, initially with ARA-mWB$_{base}$ correctly predicted the realization عز but, with ARA-mWB$_{80}$ and then ARA-mWB$_{160}$, it predicts عن: this prediction error may be explained by the underrepresentation of the word عز in the manuscript where the extremely common preposition عن occurs on many occasions.

Comments on predictions for MS.ARA.1926 and on the evolution and improvement of the model, batch after batch, highlight some blindspots and limits of assessments based on CER. The *Character Error Rate* does not take into account the distribution of the errors within pages of each manuscript. Therefore, the mean CER of MS.ARA.1922 is 12.24% but Figure 7 shows that 50% of the lines have a CER below 10%, and the same observation can be done for all the manuscripts considered. If we look closely, it appears that the main text is largely recognized by the HTR models, and the peculiarities visible in Figure 7 concern catchwords and marginal notes. Notably, the manuscripts MS.ARA.1925, MS.ARA.1926, MS.ARA.1929, MS.ARA.1944 and MS.ARA.1957 reach CER of 0% for some lines. You will find below some examples of predictions in Table 9.

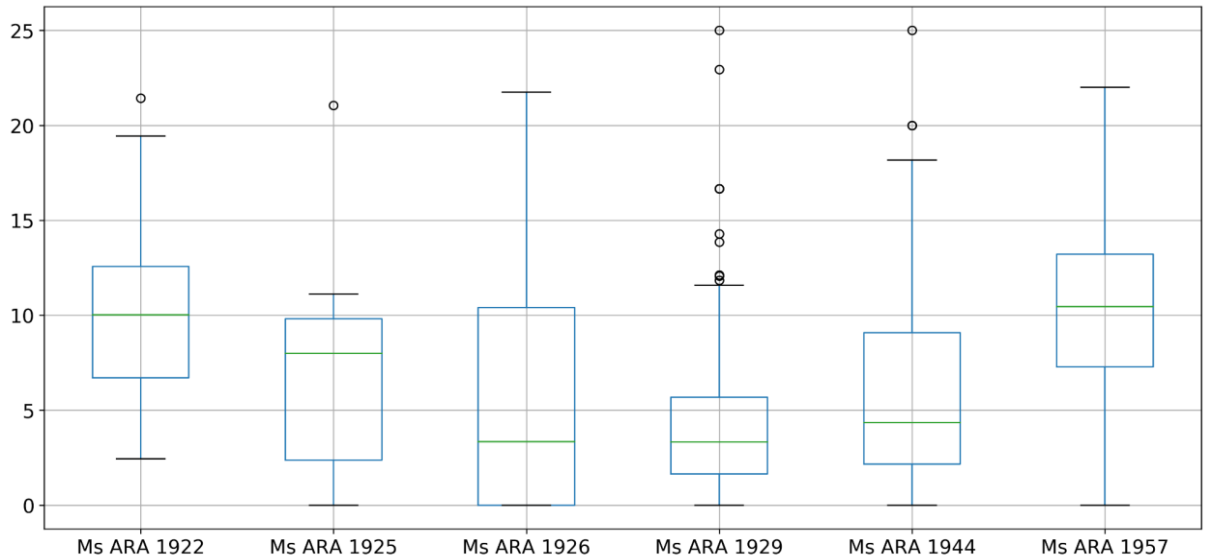

Figure 7. Distribution of CER for each manuscript

For each manuscript, an example of prediction from each final model is given in Table 9.

| MS.ARA.1922 CER: 10.24% | الرابع عشر فيما يستدل به على ارحام النسا والعواقد |
|---|---|
| GT | الرابع عشر فيما يستدل به على ارحام النسا والعواقد |
| Prediction (ARA-mWB$_{1922-160}$) | الرابع عشر فيما يبيتدل به على ارحام النساو العواقر |

| MS.ARA.1925 CER: 10% | الصالح والقطب الناصم |
|---|---|
| GT | الصالح والقطب الناصم |
| Prediction (ARA-mWB$_{1925-120}$) | الصالح والقطب الناحو |

| MS.ARA.1926 CER: 0% | 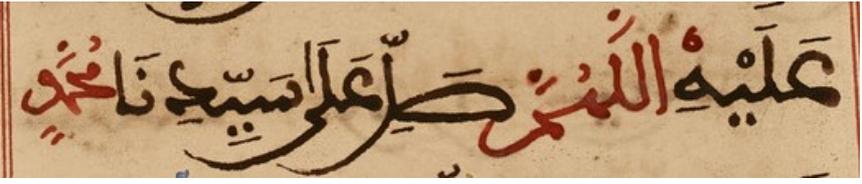 |
|---|---|
| GT | عليه اللهم صل على سيدنا محمد |
| Prediction (ARA-mWB$_{1926\text{-}160}$) | عليه اللهم صل على سيدنا محمد |

| MS.ARA.1929 CER: 2.53% | 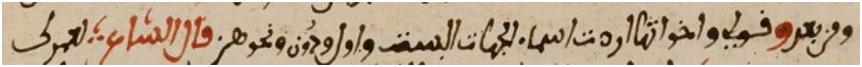 |
|---|---|
| GT | ومن بعد وقولي واخواتها اردت اسماء الجهات الست واول ودون ونحوهن قال الشاعر لعمرك |
| Prediction (ARA-mWB$_{1929\text{-}160}$) | ومن بعد وقولي واخواتها اردت اسماء الجهات الست واول ودون ونحوه قال الشاع لعمرك |

| MS.ARA.1944 CER: 1.85% | 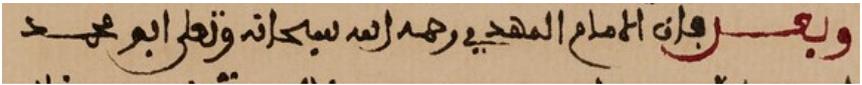 |
|---|---|
| GT | وبعد فان الامام المهدي رحمه الله سبحانه وتعلى ابو محمد |
| Prediction (ARA-mWB$_{1944\text{-}160}$) | وبعل فان الامام المهدي رحمه الله سبحانه وتعلى ابو محمد |

| MS.ARA.1957 CER: 4.47% | 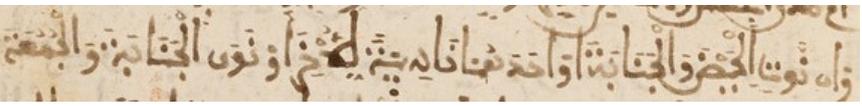 |
|---|---|
| GT | وان نوت الحيض والجنابة او احدهما ناسية للاخر أو نوى الجنابة والجمعة |
| Prediction (ARA-mWB$_{1957\text{-}160}$) | وان نوت الحض والجنابة او أحدهما ناسية للاخر او نوى الجنابة والجمعة |

Table 9. Examples of predictions on manuscripts evaluated with the last model specialized approach mWB

These results confirm that it is possible to fine-tune/specialize quickly, at lower costs, HTR models on Arabic Maghribī scripts. Most of the prediction errors are consistent with the observations we made when we first assessed the RASAM dataset (Vidal-Gorène, *et al* [2021b]: 279). The final *dāl* curved and downward is predicted as a *lām* like in the example of بعد in the MS.ARA.1944 or as a *rāʾ* in the MS.ARA.1922 for the last word of the line العواقد. Some characters, according to their position within the word and how the scribe realizes them, tend to be less well deciphered: it happened for example in MS.ARA.1957 where the *yāʾ* in الحيض, copied below the article and the *ḥāʾ*, has not been predicted. It is also the case in the MS.ARA.1925 where the final *mīm* in الناصم has been confused with a *wāw*. At the margin, we also notice some erroneous hyphenation between words as in the example of MS.ARA.1922 (النسا والعواقد instead of النساو العواقر).

It is also worth pointing out that some prediction errors are actually not errors. As the CER calculation is based on the comparison between the predicted text and the ground truth, it is affected by any deviation from this transcription. Thus, if the transcriber follows a certain specification for transcribing the text, modernizes the spelling or introduces typos himself, a "good" prediction that corresponds well to the expected result will be considered as wrong. Measuring the error rate in the present results therefore involves an experimental bias. In the case of the MS.ARA.1957 example, the transcriber did not perform the *hamza* elicited at the *alif* of أحدهما when it was present in the manuscript due to the specifications. The prediction is therefore not, in this sense, incorrect.

In the end, it is clear that an architecture based on a meta-word approach enables to overcome one of the main challenges with regard to the recognition of the handwritten Arabic scripts in manuscripts: the question of the diacritics, in particular the *iʿjām* or the phonetic distinctions of consonants. We usually observe, when it comes to the diacritics of consonants that are subscribed or underwritten, that they are placed irregularly and thus are shifted a few millimeters to the right and/or left and placed on or under another letter (Vidal-Gorène, *et al* [2021b]: 278). This problem only arises at the margin with a meta-word based approach: in the examples of Table 9, we find only one occurrence of this type of error (MS.ARA.1922 for ببيتدل) which can be explained, logically enough, by the particular realization of the *yāʾ* by the scribe as well as the presence of noise on the paper used as a support for the writing (a slightly red stain under the letter *sīn*, which probably led the model to see it as a *bāʾ*)

Overall, some additional explanations of the CER results for each of the six manuscripts considered can be provided. The CER of the MS.ARA.1944 (5.33%) is a relatively straightforward one to explain. As a manuscript dealing with history, MS.ARA.1944 shares a similar lexicon with RASAM. On top of that, its layout is simple and its writing is also close to MS.ARA.1977 included in the RASAM dataset. The best CER result is reached for MS.ARA.1929 (4.66%), whose script is also similar to MS.ARA.1977. Offering nonetheless new lexicons, since it is a grammar text, the consistency of the vocabulary and the syntax might explain the result.

New vocabulary and high variations within a single manuscript in terms of words and syntax might explain why the MS.ARA.1922 has a CER of 12.24%, the highest of the dataset. MS.ARA.1922 comprises a treaty of erotology which entails non common verbs or nouns. Moreover, this specificity is combined with a type of Maghribī script not yet encountered in the annotated manuscripts. As for MS.ARA.1957, which has the second worst CER with 10.41% (a result that nevertheless allows to work with the predicted text), we can assume that the fact that the manuscript was entirely vocalized, whereas we did not yet took into account vocalizations in our transcription, might be a factor of explanation.

To gain a better understanding of the results, we also tried to analyze pages where the CER was higher than expected (exceeding 15%). A closer look to these pages shows that pages where more mistakes occur include marginal or interlinear notes, catchwords and color.

# III. Conclusions

In this paper, we intended to address the question of handwritten text recognition for Arabic manuscripts. In particular, we aimed at assessing the use of the RASAM dataset, which we previously created, for the analysis of other manuscripts in Arabic Maghribī scripts. This investigation, led within the specific context of an important lack of annotated data, enabled us to highlight the relevance of a new neural approach for Arabic scripts, combined with an iterative process of transcription. We have found that a dedicated word-based approach, nay a meta-word based architecture (named in the course of this paper ARA-mWB architecture), exceeds state-of-the-art generic character-based approaches in such a context. Although these show satisfactory results, the word-based approach is less sensitive to the noise that may be present in the dataset (e.g. typos) and faster in training and in overcoming issues related to these scripts, in particular the separations between words and the random distribution of diacritical points (*iʿjām*).

The core objective of this contribution was also to evaluate the versatility of the RASAM dataset for the researcher who would like to use it on his own corpus in Maghribī scripts[44]. Our experiments showed a rapid specialization of the model after five transcribed pages and an accuracy between 87% (for the worst results) and 95% after ten transcribed pages. The fine-tuning approach, as well as the contribution of diversified data to our meta-word based architecture, help to solve some of the issues in terms of HTR of Arabic Maghribī script manuscripts. The challenge raised by colored inks is certainly partly overcome by this approach, but it must be stressed that too clear inks may still be problematic. All in all, it appears that fine-tuning, associated with a meta-word architecture, allows for exploitable results for limited time and data.

The evaluated model does not take into account the issue of vocalization: in the current state, the vocalization signs, that are sometimes mistaken for diacritics, introduce noise that needs to be handled. This is why we believe that one of the development steps to be considered would be to devise a model that can process vocalization signs. We are currently exploring this objective. The other target we are working on is the evaluation of the porting of the RASAM dataset to other types of scripts than Khaṭṭ Maghribī. We are exploring the opportunities offered by transfer learning in the development of specialised models for oriental scripts in particular.

---

[44] A similar experiment in real situation was during the first trimester 2022 by the GIS MOMM and the BULAC, within the frame of a new hackathon for the transcription of 15 manuscripts in Maghribī scripts from the BULAC. The results achieved confirm the models very quick specialization with only 10 pages in training.
A. Perrier, C. Vidal-Gorène, "Le développement du dataset Rasam : deux expériences de transcription collective à la BULAC", presentation at the Spring School *Les manuscrits maghrébins et les humanités numériques*, Paris, May 2nd, 2022, https://medihal.archives-ouvertes.fr/hal-03660889 [video].

In this article, we have limited ourselves to Arabic manuscripts copied in Maghribī script. Therefore, we now wish to explore the practicality of this "specialized" model for Oriental scripts in order to evaluate possibilities and benefits of transfer learning. The first tests we have been able to carry out so far indicate that transfer learning is indeed observable, but these are preliminary results. There is no doubt in our mind that deep learning technology combined with HTR offers unique possibilities for Arabic manuscripts. Our recent developments and achieved results within the scope of this research demonstrate that technologies are now robust enough to consider massive processing of Arabic manuscripts, within a defined and delimited research project. Several applications of text search or data mining, pre-editing and enrichment of the text, obtained through HTR, can now be implemented to enrich accessibility of manuscripts in digital libraries.

## Acknowledgments

This work was carried out within the scope of the Research Consortium Middle-East and Muslim Worlds (GIS MOMM) initiatives for Arabic HTR. We would also like to thank the BULAC library for providing us HD scans of manuscripts and Aliénor Decours-Perez for the translation and proofreading of this paper.

## Bibliography and Links

https://digitalorientalist.com/2021/09/24/a-study-on-the-accuracy-of-low-cost-user-friendly-ocr-systems-for-arabic-part-2/

**Maghribī script history bibliography**

**Digital philology bibliography**

**Links to dataset**

BADAM: https://zenodo.org/record/3274428#.YW_xJhrMI2w
RASAM : https://github.com/calfa-co/rasam-dataset

RASM: https://bl.iro.bl.uk/concern/datasets/f866aefa-b025-4675-b37d-44647649ba71?locale=en

**Links to websites of institutions and projects**

BINA : https://bina.bulac.fr/
Bibliothèque Nationale du Royaume du Maroc: http://bnm.bnrm.ma:86/
Centre d'Etudes Maghrébines en Algérie: https://aimsnorthafrica.org/cema/
HTR project of the British library: https://www.bl.uk/projects/arabic-htr
Fondation du Roi Abdul-Aziz Al Saoud: http://www.fondation.org.ma/web/accueil
Tunisian Office of Center for Middle East Studies: https://cmestunisia.fas.harvard.edu/

**Links to softwares, platforms and tools**

ABBYY Fine Reader PDF: https://pdf.abbyy.com/pricing/
Calfa Vision platform: https://vision.calfa.fr
Convertio: https://convertio.co/
Free Online OCR :https://www.newocr.com/
Gold/Sakhr: http://www.sakhr.com/index.php/en/solutions/ocr
i2 OCR: https://www.i2ocr.com/free-online-arabic-ocr
OCR Convert: (https://www.ocrconvert.com/arabic-ocr
OCR Space: https://ocr.space/
Online Convert Free: https://onlineconvertfree.com/ocr/arabic/
Sotoor: https://rdi-eg.ai/optical-character-recognition/